\documentclass[letterpaper, 10 pt, conference]{ieeeconf}  

\usepackage{amsmath,amsfonts}
\usepackage{algorithmic}
\usepackage{array}
\usepackage[caption=false,font=normalsize,labelfont=sf,textfont=sf]{subfig}
\let\labelindent\relax
\usepackage{enumitem}
\usepackage{booktabs} 
\usepackage{multirow} 
\usepackage{tabularx} 
\usepackage{textcomp}
\usepackage{stfloats}
\usepackage{url}
\usepackage{verbatim}
\usepackage{color}
\usepackage{soul}
\usepackage{xcolor}
\usepackage{graphicx}
\usepackage{makecell} 
\def\BibTeX{{\rm B\kern-.05em{\sc i\kern-.025em b}\kern-.08em
    T\kern-.1667em\lower.7ex\hbox{E}\kern-.125emX}}
\usepackage{balance}

\IEEEoverridecommandlockouts                              

\title{
Pose-Aware Modeling to Mitigate Pose-Related Artifacts in Tactile Gloves
}

\author{Tianhong Catherine Yu$^{1,*}$, Ziyi Kou$^{2}$, Mia Huang$^{2}$, Taylor Niehues$^{2}$, \\
Yiyue Luo$^{3}$, Li Guan$^{2}$, and Dingtian Zhang$^{2}$\vspace{-10pt}
\thanks{* Work done during an internship at Meta Reality Labs}
\thanks{$^{1}$ Cornell Univeristy, $^{2}$Meta Reality Labs, and $^{3}$University of Washington}%
}

\begin{document}
\newcommand{\etal}{et al.~}
\definecolor{CAT-comment}{rgb}{0.95, 0.2, 0.8}
\definecolor{tactile-only-estimate}{RGB}{160,160,160}
\definecolor{pose-aware-estimate}{RGB}{243, 206, 70}
\definecolor{pink}{RGB}{226,145,185}
\newcommand{\hlgray}[1]{{\sethlcolor{tactile-only-estimate}\hl{#1}}}
\newcommand{\hlyellow}[1]{{\sethlcolor{pose-aware-estimate}\hl{#1}}}
\newcommand{\hlpink}[1]{{\sethlcolor{pink}\hl{#1}}}
\newcommand{\CAT}[1]{{\color{CAT-comment} #1}}

\maketitle
\thispagestyle{empty}
\pagestyle{empty}

\begin{abstract}

Tactile gloves digitize contact and force during hand-object interactions, enabling robotics applications in dexterous manipulation, teleoperation, and learning from demonstration. To preserve hand dexterity and capture the nuances of natural interactions, these gloves and the integrated tactile sensors are designed to be soft, flexible, and comfortable. However, such flexible sensors are sensitive not only to contact forces but also unavoidably to hand pose changes, resulting in pose-related artifacts (PRAs). PRAs are especially problematic in the low-force range, resulting in misdetections or late-onset detections of contact, which raises the minimum detectable force (MDF) of the glove. In this work, we characterize the PRAs in relation to pose and force. Building on these insights, we introduce a glove-agnostic algorithmic framework that leverages hand pose information, which is increasingly available, to mitigate PRAs without glove modifications. Our pose-aware force estimation model augments tactile-to-force pipelines with a residual prediction branch that explicitly accounts for pose-induced sensor deformations. We validate our approach across 3 glove designs and 15 users, reducing MDF by 10.4\%, 12.2\%, and 18.3\%, with consistent improvements across all evaluated metrics. This method provides a practical path to improving the usability of tactile gloves in data collection and diverse robotic applications.

\end{abstract}

\section{INTRODUCTION}
\begin{figure}[!t]
    \centering
    \includegraphics[width=\linewidth]{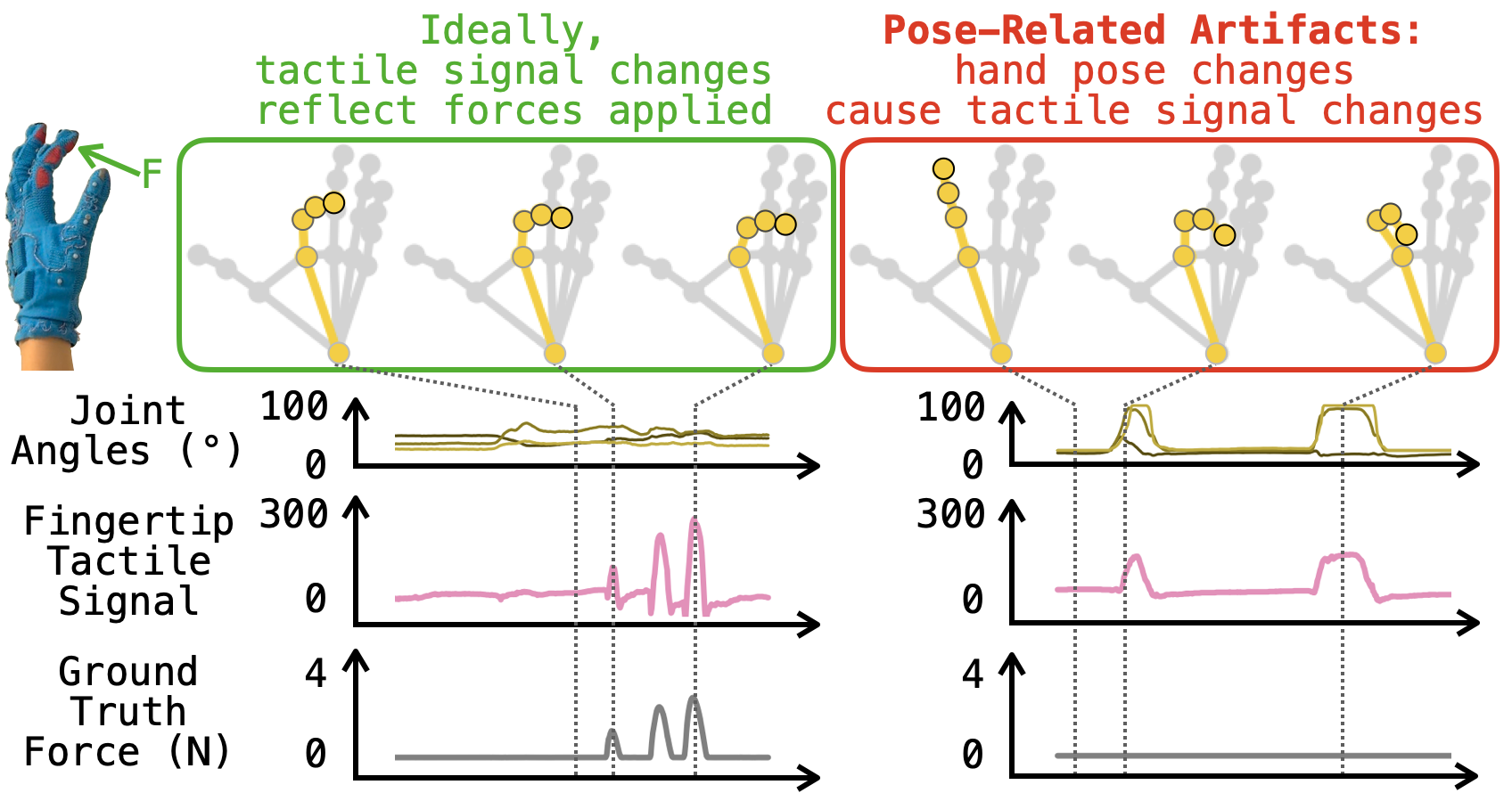}
    \caption{In tactile sensing gloves (such as Glove 2 shown at the top left), changes in \hlpink{tactile} signals should ideally reflect \hlgray{forces} applied to the hand. However, because the tactile sensors are flexible and stretchable, they are also sensitive to strain from \hlyellow{hand pose} changes. As illustrated in the red box, changing hand pose can produce tactile signal changes even without external contact. We refer to these as \textbf{pose-related artifacts (PRAs)}, whose magnitudes can be comparable to those produced by light touch, leading to misinterpretation of tactile signals.
    }
    \label{fig:teaser}
    \vspace{-10pt}
\end{figure}

Capturing tactile information during natural hand-object interactions is critical for understanding manipulation and training dexterous robots.
Soft tactile gloves have emerged as a promising solution, using flexible sensors that convert mechanical deformations into electrical signals~\cite{STAG,knit-human-env,resistive-pose-force-glove} to support learning from demonstration, teleoperation, and multi-modal sensing~\cite{VTDexManip, PiMForce, visual-tactile-joint}.
However, a longstanding challenge in flexible and soft sensor design is the intrinsic coupling between force and strain~\cite{soft_strain_pressure_survey}.
Soft sensors are sensitive to all forms of deformation, thus when integrated into gloves, they respond to both contact force (which is what tactile gloves should capture) and the wearer’s own hand pose changes, as shown in Fig.~\ref{fig:teaser}.
While this broad sensitivity enables rich sensing of hand-object interactions and hand kinematics, it also complicates isolating signals arising from true contact events versus those produced by changes in hand pose.

\textbf{We define pose-related artifacts (PRAs) as changes in tactile sensor signals caused by hand poses, rather than by contact forces.}
PRA is a fundamental challenge observed across sensor and glove designs in prior work~\cite{fits-like-a-flex-glove}, commercial systems~\cite{PPS-practical-uses}, and our own experiments.
This consistency indicates that PRAs arise from inherent force-strain coupling rather than specific hardware implementations.
Fig.~\ref{fig:teaser} illustrates example PRAs from our collected dataset (detailed in Sec.~\ref{sec:data-collection}): bending a finger in air produces a tactile signal change comparable to that of a 1.1N (or 112.2gF) tap, even at the distal phalanx (fingertip), which does not directly bend along with finger joints.

We focus our analysis and modeling on index fingertip tactile signals, as fingertips are the primary contact interface for grasping and manipulation.
Critically, fingertip sensors present a revealing case for understanding PRAs: despite being located at the distal phalanx rather than directly at joint creases, they still exhibit substantial PRAs.
This focused scope allows controlled, high-quality force measurements necessary for systematic characterization.

PRAs degrade tactile signal quality in two key ways: (1) as fingers bend without any external load, tactile readings can increase similarly to a forceful interaction, making reliable touch detection challenging; and (2) when loaded, pose-induced strain modulates the force-tactile relationship~\cite{tactile-glove-strain-correct, curvature-aware-calib}.
As a result, the same tactile reading may correspond to various contact force depending on hand pose, complicating downstream learning and manipulation tasks. 

The effects of PRAs are the most pronounced in the low-force range, critical for both human (e.g., average force during touchscreen taps is 0.5N~\cite{touchscreen-force}, or 51gF) and robotic dexterity.
Prior work has excluded forces less than 0.2N (or 20.4gF) in supervised force estimation due to glove noise~\cite{PiMForce} and found that the touch detection threshold significantly affected policy training outcomes~\cite{VTDexManip}.
Commercial Pressure Profile System (PPS) glove recommends pose-specific calibration to remove residuals from flexing~\cite{PPS-practical-uses}, but frequent calibration is impractical for natural interactions. 
Novel hardware solutions require glove redesign and specialized materials/structures~\cite{tactile-glove-strain-correct, strain-insensitive-elastomer-glove}, potentially compromising the wearability of tactile gloves.

\begin{figure}[t]
    \centering
    \includegraphics[width=\linewidth]{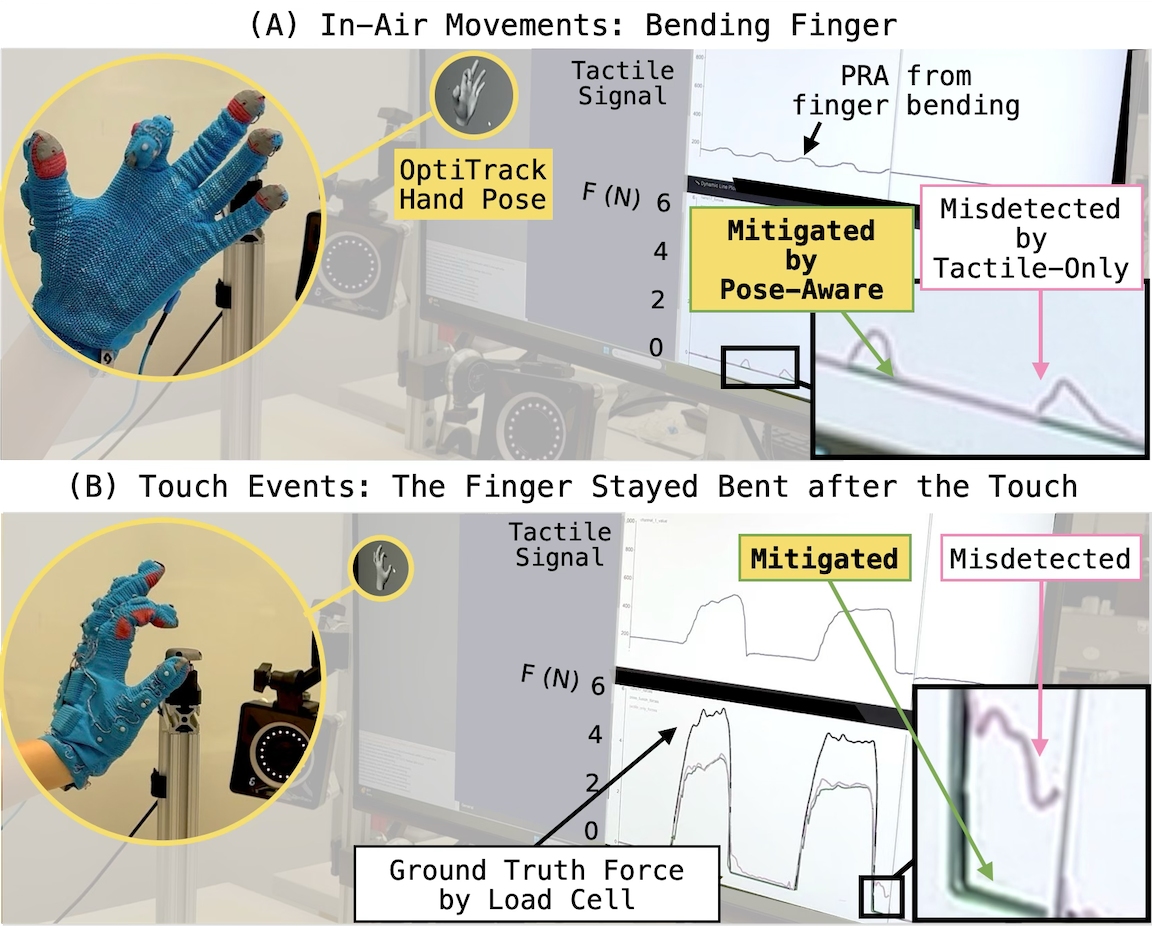}
    \caption{
    Examples of real-time PRAs mitigation for Glove 2 using our pose-aware force estimation model: (A) in-air movements and (B) touch events. Videos are included in the supplementary materials.}
    \label{fig:real-time}
    \vspace{-12pt}
\end{figure}

In this paper, we \textbf{investigate algorithmically mitigating PRAs based on the hypothesis that explicitly incorporating the hand pose modality improves force estimation by accounting for pose-induced sensor deformations}. 
The exact pose-induced strain depends on sensor mechanics, hand shapes, glove fits, etc., and can only be measured with sensor modifications. 
However, hand pose, the kinematic source of these strain variations, is increasingly available in robotic systems through motion capture~\cite{optitrack}, egocentric vision~\cite{egocentric-hand, hamer}, or wearable sensors~\cite{data-gloves-review}.
To the best of our knowledge, we are the first to use an algorithmic approach, requiring no glove hardware modifications and can be retrofitted to existing tactile gloves, to mitigate PRAs in tactile gloves.
Fig.~\ref{fig:real-time} shows our pose-aware mitigation in real time.
Our contributions are as follows:
\begin{itemize}[leftmargin=*, nosep]
    \item \textbf{Characterizations of PRAs}: We characterize PRAs in tactile gloves, providing insights critical for modeling force estimation and hand-object interaction.
    \item \textbf{Pose-Aware Force Estimation Model}: We propose a multimodal model that fuses the additional hand pose modality to mitigate PRAs through residual learning, which augments tactile-to-force estimation with an additional branch that predicts and corrects pose-induced error.
    \item \textbf{Multi-Glove Validation}: We validate our method on data collected from 3 different tactile gloves and 15 users, covering a variety of pose-force combinations. For 3 gloves, the minimum detectable force (MDF) was reduced by 10.4\%, 12.2\%, and 18.3\%, respectively, with consistent improvements observed across all other evaluation metrics.

\end{itemize}

\section{RELATED WORK}
\textbf{High-quality human tactile data.}
Collecting large-scale human-object manipulation datasets with high-quality tactile data remains challenging despite the shown effectiveness of visuotactile policies~\cite{VTDexManip}.
Among recent efforts in sensing contacts through various modalities~\cite{pressurevision++, Egopressure}, tactile gloves offer the most direct and natural approach to collecting manipulation datasets from human demonstrations~\cite{VTDexManip, tactile-multi-task-embeddings, grasp-like-humans}.
The quality of tactile data from gloves directly impacts downstream learning performance~\cite{VTDexManip, PiMForce}, motivating the need for accurate touch detection and force estimation from tactile gloves.
Achieving this high fidelity is complicated by inherent sensor limitations.

\textbf{Decoupling force from strain in flexible and soft sensors.}
A fundamental challenge in soft tactile sensors is decoupling force from strain-induced artifacts~\cite{soft_strain_pressure_survey}. 
Jiang~\etal developed a novel sensor for tactile gloves that separately measures strain and normal force~\cite{tactile-glove-strain-correct}. 
Their approach uses the sensor-measured strain to correct the estimated force by fitting a corrected force-tactile response curve to reconstruct the grasped deformable objects.
In contrast, we address PRAs with a multimodal approach that does not require any modifications to the glove design.
Our multimodal approach leverages strain-related information from an additional modality, hand pose, to correct for artifacts and decouple force from sensor signals.
Analogously, Dong~\textit{et al.} fused EIT tactile signals with 3D point cloud scans, using the geometric data to compensate for deformation-induced strain artifacts~\cite{EIT-deformable}.
In this work, we use hand pose, which encodes pose-induced strain, to provide additional information for correcting tactile-only force estimation.

\textbf{Residual learning in robotics} refers to learning a task-specific correction to a baseline robotics prediction/model (not to be confused with residual network)~\cite{residual-dynamics-learning}.
For visuotactile policy learning, Zhu~\textit{et al.} predicted a residual rotation action from an image that shows the object pose to augment and correct the tactile input~\cite{rotation-residual}.
Similarly, residual learning has emerged as an effective approach to bridge sim-to-real gaps.
The learned residual term models error between simulated and physical systems to enable more accurate soft robotic control~\cite{soft-robot-residual-physics} or calibrate visuotactile sensors~\cite{TensorTouch}.
In our work, we use residual learning to learn a residual term that specifically models pose-induced artifacts, enabling more accurate force estimation.
Unlike sensor-based approaches that require glove redesign, our method can be retrofitted to any existing system with hand pose information, making it an immediately deployable and practical solution.

\section{EMPIRICAL OBSERVATIONS \& IMPLICATIONS}\label{sec:pra-understanding}
This paper focuses on PRAs in fingertip tactile signals, as fingertips are the primary interface for tactile interactions.
For practical reasons related to force ground truth collection, we center our study on the index fingertip.
The proposed characterization and modeling approach, however, is general and can be extended to other hand locations in future work.

In an ideal scenario, tactile signals would thus reflect only contact forces, independent of hand configuration.
However, in practical tactile glove systems, this orthogonality breaks down. 
Due to the glove’s physical construction and the force-strain coupling of its sensors, changes in hand pose induce deformations in the tactile sensors, resulting in PRAs in the tactile signal, which can confound the tactile interpretation.

Through empirical observations, we identified three key properties of PRAs that challenge reliable tactile interpretation.
These empirical observations motivated the design of our data collection protocol and, more importantly, informed a set of concrete design principles for pose-aware tactile glove systems.
These principles directly guide the architecture of our pose-aware force estimation framework.

\begin{figure}[t]
    \centering
    \includegraphics[width=\linewidth]{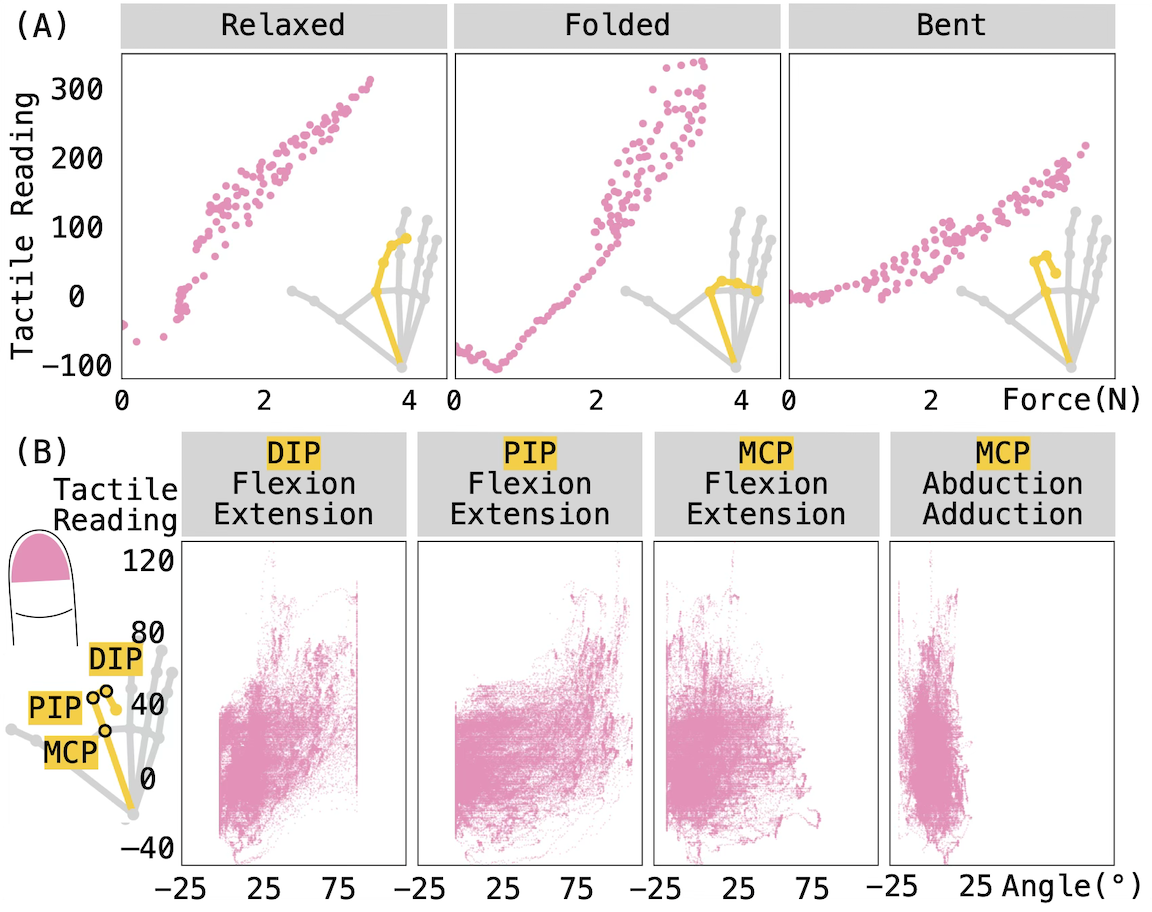}
    \caption{
    (A) Different index finger poses result in different sensor responses to force applied to the tactile gloves.
    (B) When analyzing changes in tactile readings due to in-air pose changes, the flexion/extension angle of joints closer to the fingertip shows a stronger correlation with the sensor outputs.}
    \label{fig:observations}
    \vspace{-12pt}
\end{figure}
\textbf{Observation \#1: At zero load, pose changes produce tactile signal changes.}
When moving fingers in air without any contact, tactile sensors 
exhibit measurable signal changes (Fig.~\ref{fig:teaser}A). 
These zero-load artifacts can be similar in magnitude to light contact events, suggesting that pose-induced deformation alone can trigger false positive touch detection and complicate reliable touch detection at low forces.

\textbf{Implication \#1: Leverage pose for PRA mitigation.}
Tactile glove systems could explicitly use pose information to model and predict the PRA in the tactile signal, thereby improving tactile signal qualities.

\textbf{Observation \#2: Under load, hand pose modulates force-tactile sensitivity.}
When applying force with different hand poses, we observed that the same tactile reading can correspond to different force magnitudes (Fig.~\ref{fig:observations}A). 
This suggests that pose-induced strain alters the sensor’s mechanical state, thereby modifying its force response curve—a phenomenon previously observed in the context of deformable-object-induced strain~\cite{tactile-glove-strain-correct}.

\textbf{Implication \#2: Explore pose and tactile joint representation space for artifact modeling.}
Effective artifact modeling should jointly account for pose and tactile features to capture their interaction, rather than treating pose as the sole factor. This enables more accurate force estimation across diverse hand poses and loading conditions.

\textbf{Observation \#3: PRAs are primarily driven by pose changes of nearby joints, indicating spatial locality.}
Fingertip tactile sensors respond most strongly to flexion of nearby joints (DIP and PIP), compared to more distal joints (MCP) (Fig.~\ref{fig:observations}B) and other fingers. 
This suggests that PRAs exhibit spatial locality, with sensor deformation primarily caused by kinematically proximate joint motion.

\textbf{Implication \#3: Locality enables targeted correction.}
By prioritizing pose data from kinematically proximate joints (e.g., DIP, PIP, and MCP for fingertip sensors), the system can effectively correct for pose-induced artifacts.
\begin{figure}[b]
    \centering
    \vspace{-14pt}
    \includegraphics[width=\linewidth]{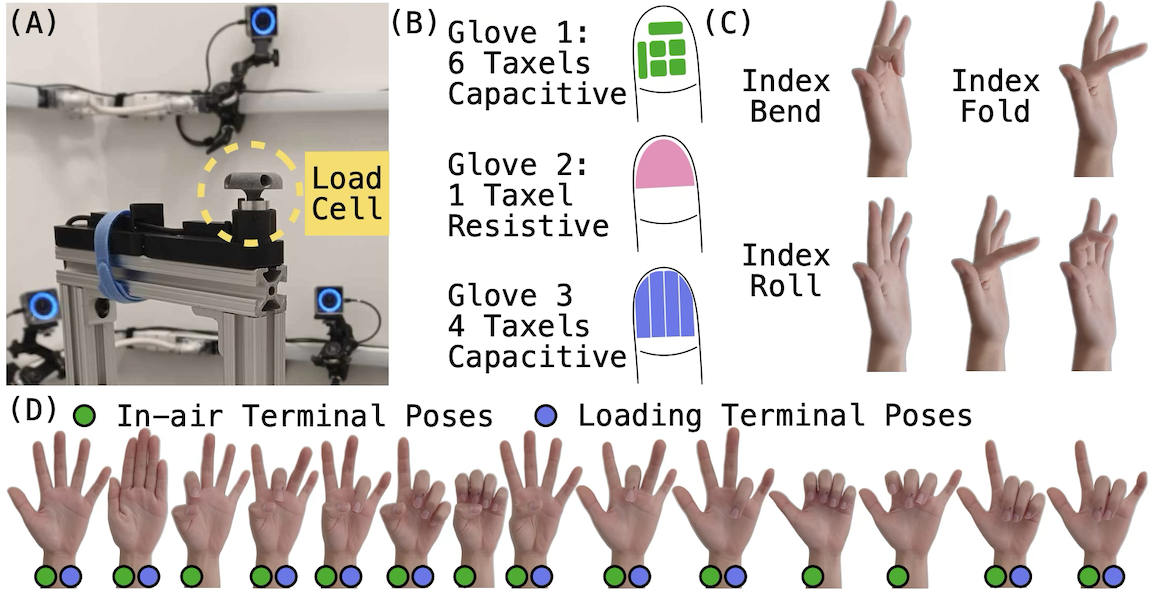}
    \caption{
    (A) Data collection setup. (B) Illustrations of the fingertip taxel layouts of the gloves used in data collection. (C) Index finger bending, folding, and rolling demonstrations. (D) Terminal poses. }
    \label{fig:data-collection}
\end{figure}
\section{DATA COLLECTION}\label{sec:data-collection}
To systematically characterize PRAs in fingertip tactile signals across different gloves, we collected data from 15 users (5 per glove), each assigned to one of 3 left-handed glove prototypes (including an off-the-shelf capacitive glove by PPS~\cite{PPS-glove} as Glove 1) with various sensor layouts and sensing principles, as shown in Fig.~\ref{fig:data-collection}B.
All experimental procedures were reviewed and approved by the institutional privacy and compliance board at the authors' organization, which serves as the equivalent oversight body for research ethics.
Each participant provided informed consent.

Users were selected to match glove sizing requirements. 
\textit{Hand pose} was captured using the OptiTrack motion capture system~\cite{optitrack}.
\textit{Force ground truth} was measured with a 6-axis ATI Nano17 load cell (resolution and minimal detectable force: 0.32 gF) and a custom flat end effector (Fig.~\ref{fig:data-collection}A), simulating a flat surface or object.
Users were instructed to apply force in the normal direction using their index fingertip. 
All glove, OptiTrack, and Nano17 signals were synchronized and resampled to 100~Hz.
Stage annotations were automatically recorded via a custom user interface.
Each session was structured into three stages:

\textbf{Stage 1: In-Air Zero-Load Pose Changes.} 
To capture pure PRAs without confounding effects from contact forces, users performed in-air pose changes like that in a pose tracking study. This stage includes:
\textit{(1) Terminal poses}: 14 hand poses (Fig.~\ref{fig:data-collection}D), each held for 2 seconds, 5 repetitions, randomized order.
\textit{(2) Index finger movements}: 3 dynamic movements (Fig.~\ref{fig:data-collection}C): bend (primarily involving DIP \& PIP), fold (MCP), and roll (DIP, PIP, MCP), each performed for 2 seconds, 10 repetitions, randomized.
\textit{(3) All-finger movements}: the same 3 movements performed simultaneously with all five fingers to assess potential multi-finger effects.

\textbf{Stage 2: Load-Unload Pose Changes.}
To examine tactile responses to different hand poses under varying load, such as at the beginning/end of a touch or grasp, users performed a series of tappings and pressings (instructed to be under 10N) across a range of static and dynamic poses. This stage includes:
\textit{(1) Terminal pose taps}: 3 consecutive taps at the instructed light, medium, and heavy force levels; 10 hand poses (index fingertip available for tapping, Fig.~\ref{fig:data-collection}D); 5 repetitions, randomized order (same for step \textit{(2, 3)}).
\textit{(2) Terminal pose presses}: 5-second sustained presses with varying forces within the press; same 10 poses.
\textit{(3) Index pose taps/presses}: after assuming one of 5 index finger poses (straight, small bend, large bend, small fold, large fold), 3-tap sequence or a sustained press to isolate force changes from pose changes.
\textit{(4) Index movement taps/presses}: dynamic transitions from straight to the other 4 index poses while tapping/pressing, simulating natural movements such as grasping.
\textit{(5) Free movement taps/presses}: 10 3-tap sequences and 10 presses during free finger movements for pseudo-unseen combinations of poses and forces.

\textbf{Stage 3: Loaded Pose Changes.}
To characterize PRAs during continuous loading, users performed explicit pose changes while maintaining a press, similar to actions in grasping a deformable object. This stage includes:
\textit{(1) Terminal pose transitions}: transitions among 10 static hand poses (as defined above) during a continuous press; 2 seconds per transition, 5 repetitions, randomized order.
\textit{(2) Index movement sequence}: continuous sequence of press, bend, unbend, fold, unfold, and unpress actions with the index finger, within a 6-second window; 10 repetitions.
\textit{(3) All-finger movement sequence}: same movement sequence as above, performed simultaneously with all five fingers.
\textit{(4) Free finger movements during press}: free finger movements during 10 presses, providing pseudo-unseen pose changes.

After each stage, users performed \textit{in-air free finger movements} to increase data diversity and to calibrate for potential drift in the force ground truth.
Each session yielded about 166,200 tactile frames, 545 taps, and 130 presses during 27.7 minutes of active data collection.
Each user completed 2 sessions, with the glove remounted between sessions to include within-user, cross-session variability.
Thus, for each glove, about 4.6 hours of data are collected, totaling ~13.8 hours and ~5M frames across all 15 users and 3 gloves. This design prioritizes within-user diversity of pose-force conditions over user count, which is appropriate for characterizing a sensor-level phenomenon (PRAs) rather than population-level variability.

\section{POSE-AWARE FORCE ESTIMATION FRAMEWORK}\label{sec:method}
Guided by the empirical insights from Sec.~\ref{sec:pra-understanding}, we now present our pose-aware force estimation framework. 
Our approach uses residual learning to explicitly mitigate PRAs.
The key idea is to preserve the strong causal mapping from tactile signals to force while modeling PRAs as a separate residual term predicted from hand pose and its interaction with tactile features.
In early explorations, we found that directly concatenating tactile and pose features for force estimation underperformed the tactile-only model, suggesting that the model mistakenly learned the pose-to-force association instead of the pose-to-artifact mapping.

\subsection{Problem Formulation}

Given a continuous temporal window of tactile readings $\mathbf{T}_t \in \mathbb{R}^{L \times T}$ and synchronized hand pose $\mathbf{P}_t \in \mathbb{R}^{L \times P}$, where $L$ is the window length, $T$ is the number of taxels, and $P$ is the number of joint angles, our goal is to estimate the applied normal force $\mathbf{F}_t \in \mathbb{R}$ at time $t$.
The last temporal frame of both $\mathbf{T}_t$ and $\mathbf{P}_t$ is aligned with $\mathbf{F}_t$ to avoid real-time latency.

The observed tactile signal reflects sensor deformation from both contact forces and pose changes. 
These effects are entangled: changes in hand pose can alter the mechanical state of the sensor, thereby modulating the tactile response to force (see Sec.~\ref{sec:pra-understanding}, Observation \#2). 
This entanglement precludes simple decomposition of the tactile signal into force and pose components via preprocessing or filtering.
We address this by modeling force estimation as a pose-aware correction process:
\begin{equation}
\hat{\mathbf{F}}_t = \hat{\mathbf{F}}_t^{\text{base}}(\mathbf{T}_t) - \hat{\mathbf{F}}_t^{\text{residual}}(\mathbf{T}_t, \mathbf{P}_t)
\end{equation}
where $\hat{\mathbf{F}}_t^{\text{base}}$ is the tactile-only force estimate and $\hat{\mathbf{F}}_t^{\text{residual}}$ predicts the signed residual induced by PRAs.

\begin{figure*}[t]
    \centering
    \includegraphics[width=\linewidth]{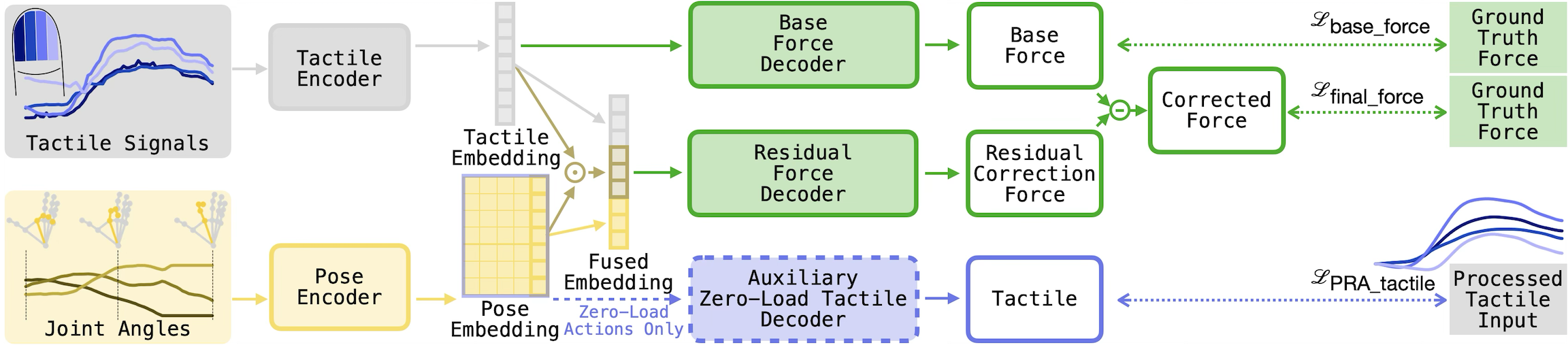}
    \caption{
    The pose-aware modeling framework takes tactile signals and joint angles as inputs, fuses pose and tactile information through a series of encoders and decoders, and outputs corrected force estimates from the main branch and processed tactile signals predicted by the auxiliary branch.}
    \label{fig:model-archi}
    \vspace{-14pt}
\end{figure*}

\subsection{Model Architecture}
Our pose-aware force estimation framework (Fig.~\ref{fig:model-archi}) is designed to flexibly incorporate both tactile and pose information, and to explicitly model their interaction.
The framework is architecturally glove-agnostic: it accommodates any tactile-to-force pipeline, as long as it produces a tactile feature representation and an estimated force.
Note the difference from being glove-independent: the tactile encoder still must be adapted and trained for a specific glove design with distinct sensing modalities and layouts.

\textbf{Tactile-Only Force Estimation Branch.} 
This branch serves as the baseline for force estimation. 
In our implementation, we instantiate this branch with a lightweight yet expressive architecture to provide a strong tactile-only baseline. 
Given $\mathbf{T}_t$, we use a \texttt{Conv1d(kernel=3, padding=1)} layer followed by \texttt{BatchNorm} and a \texttt{ReLU} activation to extract local spatial features. 
To further capture temporal dependencies over both short and long temporal horizons, we employ a multi-scale dilated temporal convolution block (3 layers of \texttt{Conv1d(kernel=3)} with dilations and paddings of 1,2, and 4)~\cite{dialted-conv} followed by a \texttt{AdaptiveMaxPool1d(1)} to produce a compact tactile feature vector, $\mathbf{T}_t^{\text{feat}} \in \mathbb{R}^{64}$.
This feature vector is decoded with 2 fully connected layers with a \texttt{ReLU} layer in between into the estimated base force, $\hat{\mathbf{F}}_t^{\text{base}}$.
By using a strong, learnable tactile-only model as the baseline, we ensure that improvements from the pose-aware correction are attributed to the explicit modeling of PRAs, rather than deficiencies in the tactile processing pipeline.

\textbf{Pose Feature Extractor Branch.} 
Given $\mathbf{P}_t$, we first apply sinusoidal encoding, $ \mathbf{P}_t^{\text{enc}} = [\sin(\mathbf{P}_t), \cos(\mathbf{P}_t)] \in \mathbb{R}^{L \times 2P}$, to handle angle periodicity and ensure continuity~\cite{angle-continuity}.
$\mathbf{P}_t^{\text{enc}}$ is transformed into an embedding of size 64 with \texttt{LayerNorm}, a fully connected layer, and a \texttt{GELU} activation.
Then, to capture the temporal changes of hand pose, we use a 2-layer bidirectional LSTM (Bi-LSTM) of hidden dimension of 32 to obtain a sequence of hidden states for all temporal frames, $\mathbf{P}_t^{\text{hidden}}$. 
For later fusion, we use the last temporal frame as pose features, $\mathbf{P}_t^{\text{feat}}  \in  \mathbb{R}^{64}$, to emphasize the latest hand pose.
For the auxiliary decoder, we use the entire sequence $\mathbf{P}_t^{\text{hidden}}$ to leverage the full temporal context.
We choose a Bi-LSTM over alternatives such as graph convolutional networks (GCNs)~\cite{MS-G3D} and PoseConv3D~\cite{PiMForce,PoseConv3D} for its lightweight yet expressive and smooth representation suitable for real-time uses.

\textbf{Feature Fusion and Residual Mitigation Module.}
To explicitly mitigate PRAs in force estimation, we fuse the tactile and pose features and use them to predict a residual correction term.
We concatenate $\mathbf{T}_t^{\text{feat}}$ and $\mathbf{P}_t^{\text{feat}}$, and their element-wise product, which captures context-dependent interactions: $\mathbf{Z}_t = \left[\, \mathbf{T}_t^{\text{feat}},\ \mathbf{P}_t^{\text{feat}},\ \mathbf{T}_t^{\text{feat}} \odot \mathbf{P}_t^{\text{feat}}\, \right] \in \mathbb{R}^{192}$.
We decode the fused $\mathbf{Z}_t$ the same as tactile-only force decoding for $\hat{\mathbf{F}}_t^{\text{residual}}$.
We calculate the final mitigated force, $\hat{\mathbf{F}}_t$ by subtracting $\hat{\mathbf{F}}_t^{\text{residual}}$ from $\hat{\mathbf{F}}_t^{\text{base}}$.

\textbf{Auxiliary Zero-Load PRA Tactile Decoding Branch.}
To further regularize the model and encourage explicit disentanglement of pose and force effects, we introduce an auxiliary decoder, with the same architecture as force decoders, dedicated to modeling PRAs in the tactile signal under zero-load conditions to decode $\mathbf{P}_t^{\text{hidden}} \in \mathbb{R}^{L \times 64}$ into $\hat{\mathbf{T}}_t^{\text{zero\_load}} \in \mathbb{R}^{L \times T}$.
Importantly, this auxiliary decoder is only supervised during zero-load (in-air) conditions: any changes in the tactile signal can be attributed purely to PRAs, providing a reliable ground truth for supervision.
With load, the pose effect cannot be separated from the tactile signals.

\subsection{Training Objective}
The training objective for our pose-aware force estimation framework is to jointly optimize accurate force prediction and explicit mitigation of PRAs, while encouraging disentanglement between force and pose effects in the tactile signal.
Our framework employs a multi-term loss, with each term targeting a specific aspect of the model’s performance.

\textbf{Corrected Force Loss.}
The primary objective is to minimize the error between $\hat{\mathbf{F}}_t$ and $\mathbf{F}_t$. 
We use a modified mean absolute percentage error (MAPE) loss to emphasize robustness at the low-force range: 
\begin{equation}
\mathcal{L}_{\text{final\_force}} = \text{MAPE}(\hat{\mathbf{F}}, \mathbf{F}) = \frac{1}{N} \sum_{i=1}^N \frac{|\hat{F}_i - F_i|}{1 + F_i}
\end{equation}
, where $N$ is the number of training samples.
The denominator includes a base term 1 to avoid division by zero and to stabilize training for small force values.

\textbf{Base Force Loss.}  
To ensure that the tactile-only branch provides a strong baseline, we also supervise the tactile-only force estimate with the same loss function: 
\begin{equation}
\mathcal{L}_{\text{base\_force}}= \text{MAPE}(\hat{\mathbf{F}}^{base}, \mathbf{F}).
\end{equation}

\textbf{Auxiliary Zero-Load PRA Tactile Loss.}  
To explicitly encourage the model to disentangle pose-induced artifacts from force effects, we introduce an auxiliary loss during zero-load (in-air) conditions. The auxiliary decoder reconstructs the tactile signal from pose features, and the mean absolute error (MAE) loss is computed as
\begin{equation}
\mathcal{L}_{\text{PRA\_tactile}} = \frac{1}{N_0} \sum_{i \in \mathcal{I}_0} \left| \hat{T}_i - T_i \right|
\end{equation}, 
where $\mathcal{I}_0$ indexes the set of frames from zero-load stages and $N_0$ is their count.
The tactile target is preprocessed with a 1D Gaussian smoothing to remove signal noise and compute the net change within the tactile window to remove signal drifting and mounting inconsistencies.
The loss term is weighted by $\lambda$.

\textbf{Total Loss.}
The overall training objective is a weighted sum of the above terms:
\begin{equation}
\mathcal{L}_{\text{overall}} = \mathcal{L}_{\text{final\_force}} + \mathcal{L}_{\text{base\_force}} + 
\lambda \mathcal{L}_{\text{PRA\_tactile}}
\end{equation}
where the two force loss terms are equally weighted and $\lambda$ is a glove-dependent hyperparameter.

\section{RESULTS}\label{sec:experiments}

\subsection{Experimental Setup}\label{sec:experiment-setup}
\textbf{Train/Test Split.}
For each collection session (i.e., a dondoff), we split the data chronologically: the first 80\% of continuous samples for training and the last 20\% for testing, following standard practice in wearable sensing to ensure that the model is evaluated on temporally later data, which better simulates real-world uses considering temporal leakage.

\textbf{Evaluated Models.}
For each glove's 5 user dataset, we trained 2 models using all 80\% of data from all 5 users and tested on the rest 20\% of data:
(1) our proposed pose-aware model; and
(2) tactile-only model, the tactile-only-force estimation branch, which ablates the pose input serving as the baseline.
\begin{table*}[t]
\caption{Results for 3 glove types: improvements across all metrics.}
\centering
\setlength{\tabcolsep}{2pt}
\begin{tabularx}{\linewidth}{c
>{\centering\arraybackslash}X >{\centering\arraybackslash}X >{\centering\arraybackslash}X >{\centering\arraybackslash}X >{\centering\arraybackslash}X 
>{\centering\arraybackslash}X >{\centering\arraybackslash}X >{\centering\arraybackslash}X >{\centering\arraybackslash}X >{\centering\arraybackslash}X 
>{\centering\arraybackslash}X >{\centering\arraybackslash}X >{\centering\arraybackslash}X >{\centering\arraybackslash}X >{\centering\arraybackslash}X }
\toprule
\multicolumn{1}{c|}{} 
    & \multicolumn{5}{c|}{Glove 1} 
    & \multicolumn{5}{c|}{Glove 2} 
    & \multicolumn{5}{c}{Glove 3} \\
\midrule
\multicolumn{1}{c}{Model} 
    & BA$\uparrow$ & F1$\uparrow$ & MDF$\downarrow$ & MAE$\downarrow$ & $R^2$$\uparrow$
    & BA$\uparrow$ & F1$\uparrow$ & MDF$\downarrow$ & MAE$\downarrow$ & $R^2$$\uparrow$
    & BA$\uparrow$ & F1$\uparrow$ & MDF$\downarrow$ & MAE$\downarrow$ & $R^2$$\uparrow$ \\
\midrule
Tactile-Only Model
    & 98.19 & 97.73 & 41.80 & 14.83 & 0.964 
    & 97.06 & 95.98 & 48.61 & 21.70 & 0.912 
    & 97.59 & 97.23 & 33.69 & 13.84 & 0.934 \\
Pose-Aware Model
    & \textbf{98.46} & \textbf{98.13} & \textbf{37.44} & \textbf{13.54} & \textbf{0.969}  
    & \textbf{97.69} & \textbf{96.90} & \textbf{42.18} & \textbf{20.51} & \textbf{0.920} 
    & \textbf{98.17} & \textbf{97.90} & \textbf{27.52} & \textbf{12.78} & \textbf{0.943} \\
\bottomrule
\end{tabularx}
\label{tab:main-results}
\end{table*}

\begin{figure*}[t]
    \centering
    \vspace{-5pt}
    \includegraphics[width=\linewidth]{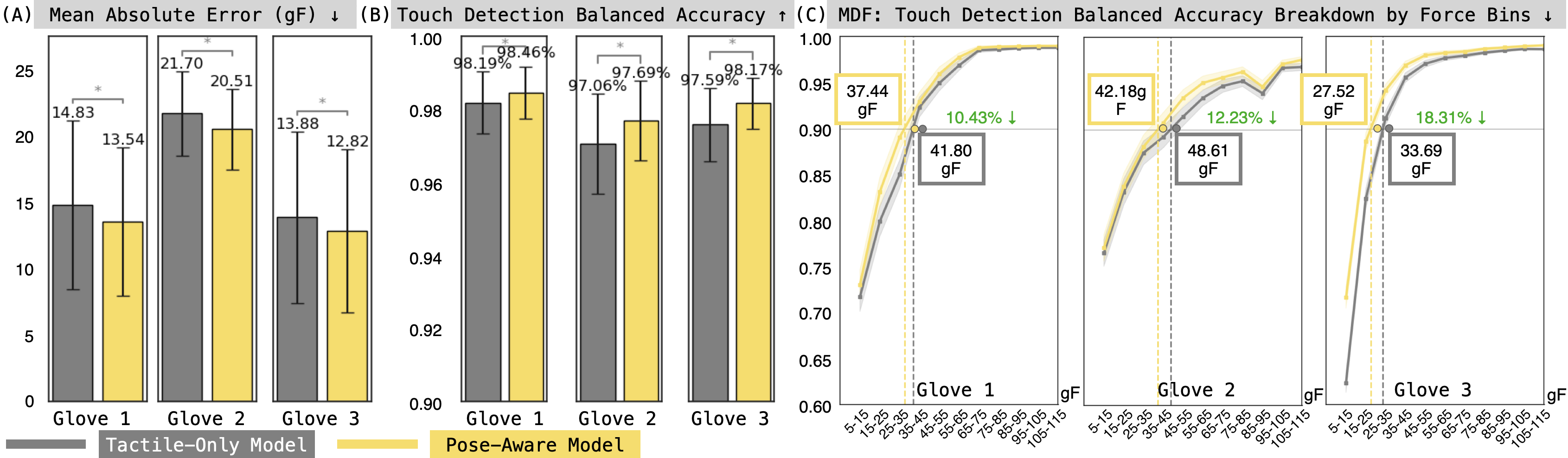}
    \caption{
    Comparison of \hlyellow{pose-aware} and \hlgray{tactile-only} models: (A) Force estimation MAE, (B) touch detection balanced accuracy, and (C) minimal detectable force (MDF, the smallest force where balanced accuracy reaches 90\%). Results show improved performance with pose-aware models. Error bars in (A, B) and shaded areas in (C) indicate 95\% confidence intervals.
    }
    \label{fig:main-results}
    \vspace{-14pt}
\end{figure*}

\textbf{Evaluation Metrics.}
To account for the drifting issue in Nano17\footnote{https://www.ati-ia.com/products/ft/ft\_models.aspx?id=Nano17}, we apply dynamic taring and select 2.5gF as the touch detection threshold as the force ground truths of all zero-load actions fall below 2.5gF.
We report the following metrics to comprehensively assess model performance:
\begin{itemize}[leftmargin=*, nosep]
    \item \textbf{Touch Detection Balanced Accuracy (BA) \& F1 Score:} The average of the true positive rate (sensitivity) for touch ($\geq2.5$gF) events and the true negative rate (specificity) for no-touch ($<2.5$gF) events (BA), along with the harmonic mean of precision and recall for the touch class (F1). Together, these metrics provide a comprehensive assessment of touch detection performance, especially in the presence of class imbalance.
    \item \textbf{Minimal Detectable Force (MDF in gF):} We define MDF as the smallest force at which the BA of touch detection reaches 90\%. BA is computed for each force bin (10gF), and the MDF corresponds to the force value where the BA–force curve first crosses 90\%. This metric quantifies the system’s sensitivity to light touch.
    \item \textbf{Mean Absolute Error (MAE in gF):} Average absolute difference between predicted and ground-truth force values, reported in grams-force (gF).
    \item \textbf{Coefficient of Determination ($R^2$):} Proportion of variance in ground-truth force explained by predictions.
\end{itemize}

\textbf{Model Inputs \& Preprocessing.}
Both tactile and pose data are processed in temporal windows of 64 frames (0.64~s at 100~Hz) with 1-frame steps, consisting of 63 historical and 1 current frames. 
For pose input, we use only the four most relevant joint angles of the index finger: DIP and PIP flexion/extension, MCP flexion/extension, and MCP abduction/adduction, motivated by our empirical findings that PRAs at the fingertip are primarily driven by these joints.
Tactile signals are normalized on a per-session basis by subtracting the channel median of zero-load stages. 
This normalization reduces the impact of session-to-session drift and donning inconsistencies, ensuring that the model focuses on dynamic changes rather than static offsets.
For real-time uses, we do a 10-second in-air movement recording as a calibration in place of zero-load stages.

\textbf{Loss Weighting.}
We conducted ablation studies to investigate the impact of the base force loss term and its weighting. 
Omitting this term led to the tactile-only branch predicting only about half the actual value, while the residual branch compensated for the remainder. 
Making the base force loss weight less than 1 resulted in degraded overall performance and less interpretable model behavior. 
Equal weighting of the two force loss terms yielded the best results.
Similarly, $\lambda$ is a hyperparameter defining the weight for the auxiliary zero-load PRA tactile loss term.
$\lambda$ is glove-dependent because each glove's tactile sensors' characteristics are different.
We selected the values for $\lambda$ by sweeping a range.

\subsection{Quantitative Results}
\begin{figure*}[t]
    \centering
    \includegraphics[width=\linewidth]{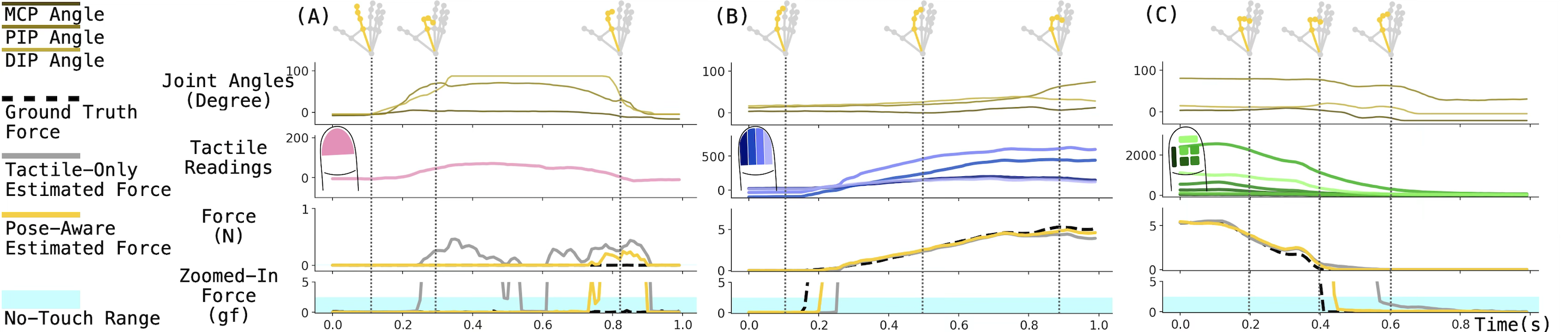}
    \caption{
    In comparison to \hlgray{tactile-only estimated force}, \hlyellow{the pose-aware estimated force} mitigates PRAs, especially for the low-force range, as shown in the bottom subplots where the blue shaded areas are the no-touch range ($<$ 2.5 gF).  
    (A) During in-air hand pose changes, tactile-only models can misdetect touches. 
    (B) During unloaded-to-loaded transitions, suppressing misdetection in tactile-only models can introduce large detection latency. 
    (C) During loaded-to-unloaded transitions, a bent finger can cause the tactile-only model to estimate lingering force even after the external load is removed.
    }
    \label{fig:qual-results}
    \vspace{-10pt}
\end{figure*}
\textbf{Improvements across All Metrics.}
Table~\ref{tab:main-results} and Fig.~\ref{fig:main-results} compare pose-aware and tactile-only models and show that adding pose consistently improves all metrics for every glove: the pose-aware model reduces MDF by 10.4\%, 12.2\%, and 18.3\%; touch detection error by 15\%, 21\%, and 24\%; and MAE by 8.7\%, 5.5\%, and 7.7\%, for Glove 1, 2, and 3, respectively.

\textbf{Notable Reductions in MDF.}
The improvements are concentrated in the critical low-force regime, where PRAs have the most impact on contact detection. Fig.~\ref{fig:main-results}C shows that the touch detection balanced accuracy of the pose-aware model consistently stays above the tactile-only model across the visualized low-force range, which reflects meaningful practical gains on PRA mitigation.

\textbf{Consistent but Varying Improvements Across Gloves.}
The variation in improvements across gloves highlights that sensor characteristics (e.g., materials and mechanics) influence the benefit of pose-aware mitigation.
While cross-glove comparisons are confounded by factors like different user populations, we observe interesting patterns containing insight into when pose information is most beneficial.
Glove 3 achieves the lowest MDF among all gloves both before (33.69 gF) and after (27.52 gF) pose-aware mitigation, while also showing the largest relative improvement (18.3\%). 
Fig.~\ref{fig:main-results}C shows that Glove 3 has a steep BA curve in the low-force range, starting with poor accuracy at ultra-low forces but rapidly improving as force increases, ultimately achieving the lowest MDF. 
This pattern suggests the sensor detects signals at these low force levels but struggles to distinguish PRA-induced signals from true forces. 
The large gap between the two MDF curves at 5-15gF and the substantial 18.3\% MDF reduction shows that much of the ultra-low force confusion for the tactile-only model stems from PRAs.
In contrast, Glove 2 shows less confusion at ultra-low forces but achieves the worst overall MDF and all other metrics, suggesting its sensors may simply be less responsive to signals in general. 
These patterns reveal a fundamental sensitivity-specificity tradeoff: without pose information, tactile-only models sacrifice low-force sensitivity to maintain specificity against PRA-induced misdetections, a limitation that the pose-aware approach mitigates.

\textbf{Statistical Analysis.}  
One-sided Wilcoxon signed-rank tests ($n=5$ users per glove, paired by user) indicate significant improvements for BA, F1, MAE, and $R^2$ ($p = 0.03125$ for all gloves and metrics). 
Effect sizes (Cohen’s $d_z$) are large, ranging from $|d_z| = 1.23$ to $2.89$, all exceeding the $0.8$ threshold for large effects and further supporting the benefit of pose-aware models.
Per-user statistical analysis was not performed for MDF due to limited data.
Our small sample size yields discrete p-values of $0.03125$, whenever all paired differences are in the same direction.
While this limits the significance assessment, the consistency of improvements across all users, gloves, and metrics provides strong evidence.

\textbf{Generalizability against ``Unseen'' Poses.}
We also evaluate the model robustness on the data during pseudo-unseen free-movement substages, where users were instructed to move their hands freely during each of the 3 stages. 
The pose-aware model improves all metrics across all gloves, and statistical significances still hold, except for MAE (14.19$\rightarrow$13.78, $\downarrow$) and $R^2$ (0.92$\rightarrow$0.93, $\uparrow$) improvements on Glove 2.
Notably, MDFs are reduced by 10.2\%, 9.6\%, and 17.0\%.
These results suggest that our pose-aware framework generalizes beyond prescribed movements, consistently mitigating PRAs even under ``unseen'' pose-force combinations.

\subsection{Qualitative Results}\label{sec:pose-and-PRA}
\begin{figure}[t]
    \centering
    \vspace{-6pt}
    \includegraphics[width=\linewidth]{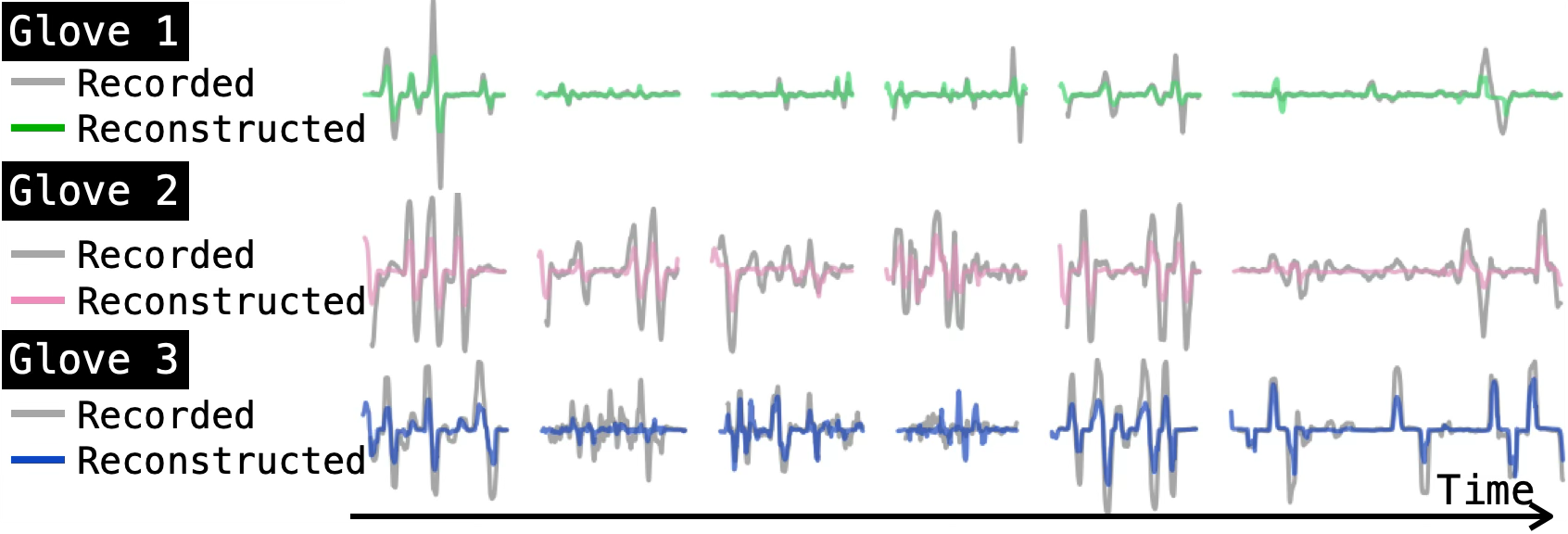}
    \caption{
    Reconstruction of tactile input (one channel from each glove) for zero-load actions.}
    \label{fig:PRA-tactile}
    \vspace{-15pt}
\end{figure}

Fig.~\ref{fig:qual-results} shows examples of how PRAs degrade tactile-only force estimation throughout the contact lifecycle and how pose-aware correction resolves these failures.
During in-air movements, PRAs cause tactile-only models to trigger false positives. 
At touch onset, conservative thresholding to suppress these false detections introduces detection latency. 
At touch offset, persistent finger bending creates lingering forces. 
The pose-aware approach mitigates these failure modes, and the latter two improve touch detection temporal sensitivities, critical for policy training~\cite{VTDexManip}.
Fig.~\ref{fig:PRA-tactile} further supports Observation~\#1: despite using a data-driven learned mapping, the clear correlation between reconstructed and recorded tactile signals for zero-load actions shows that pose alone carries predictive information for in-air PRAs.

\section{CONCLUSION and FUTURE WORK}

\textbf{Conclusion.}
This work addresses the challenge of pose-related artifacts (PRAs) in tactile glove sensing, which degrade touch detection and force estimation, especially at low forces crucial for robotics applications. 
By characterizing PRAs across multiple glove designs and users, we detail observations and insights for understanding and mitigating PRAs. 
Our pose-aware force estimation framework leverages hand pose information, now common in robotic systems, to algorithmically mitigate PRAs without glove hardware changes. 
This glove-agnostic, easily deployable approach advances robustness in tactile gloves, supporting broader applications in robotics and human-robot interaction.

\textbf{Future Work.}
Extending our method beyond index fingertip to full-hand sensing remains important future work for understanding how PRA characteristics vary across hand locations (e.g., palm, joint creases). So does zero-shot transfer to an unseen sensor or glove without retraining.
Further, our method relies on high-precision, marker-based motion capture.
As more portable and wearable systems (e.g., egocentric vision) become available, it is critical to understand how pose tracking quality affects PRA mitigation and downstream application performance.
Finally, our evaluation uses a single type of flat end effector to provide high-fidelity force ground truth, which does not capture the full diversity of real-world manipulation tasks and object interactions. 
Future work should expand to dynamic manipulations with instrumented objects in naturalistic environments, including in-the-wild data collection.

\addtolength{\textheight}{-12cm}   





\bibliographystyle{bib/IEEEtran}
\bibliography{bib/ref}

\end{document}